\documentclass[compsoc]{IEEEtran}
\usepackage{amsmath,amsfonts}
\usepackage{algorithmic}
\usepackage{array}
\usepackage[caption=false,font=normalsize,labelfont=sf,textfont=sf]{subfig}
\usepackage{textcomp}
\usepackage{stfloats}
\usepackage{url}
\usepackage{verbatim}
\usepackage{graphicx}
\usepackage{cite}
\usepackage{multicol}
\usepackage{multirow}
\usepackage{xcolor,colortbl}
\usepackage{xspace}
\usepackage{enumitem}
\usepackage{makecell}
\usepackage{booktabs}
\usepackage{hyperref}
\usepackage{algorithm2e}
\usepackage{mathtools} 
\usepackage{amssymb}
\usepackage{amsmath}
\usepackage{ulem} 

\renewcommand{\emph}{\textit}

\newcommand{\autorefapp}[1]{\hyperref[#1]{Appendix~\ref*{#1}}}

\definecolor{applepinknormal}{RGB}{255, 55, 95}
\newcommand{\revise}[1]{{#1}}

\DeclareMathOperator*{\argmax}{arg\,max} 

\definecolor{cbblue}{RGB}{0, 114, 178}
\definecolor{cbgreen}{RGB}{0, 158, 115}
\definecolor{cbred}{RGB}{213, 94, 0}
\definecolor{cbbrown}{RGB}{124, 65, 28}

\newcommand{\blue}[1]{{\color{cbblue}#1}}
\newcommand{\red}[1]{{\color{cbred}#1}}
\newcommand{\green}[1]{{\color{cbgreen}#1}}
\newcommand{\brown}[1]{{\color{cbbrown}#1}}

\def\IVMa{IVM$_{A}$\xspace}
\def\AB{A\&B\xspace} 

\def\pplus{$\mathcal{P}^+$}
\def\pentire{$\mathcal{P}^*$}
\def\pminus{$\mathcal{P}^-$}

\renewcommand{\paragraph}[1]{
\vspace{4pt}
\noindent
\textbf{#1}.
}

\begin{document}

\title{Measuring the Validity of \\ Clustering Validation Datasets}

\author{Hyeon Jeon, Micha\"el Aupetit, DongHwa Shin, Aeri Cho, Seokhyeon Park, and Jinwook Seo
\thanks{Hyeon Jeon, Aeri Cho, Seokhyeon Park, and Jinwook Seo are with Seoul National University. e-mail: \{hj, archo, shpark\}@hcil.snu.ac.kr, jseo@snu.ac.kr}
\thanks{Micha\"el Aupetit is with Qatar Computing Research Institute, Hamad Bin
Khalifa University. e-mail: maupetit@hbku.edu.qa}
\thanks{DongHwa Shin is with Kwangwoon University. e-mail: dhshin@kw.ac.kr.}
\thanks{Jinwook Seo is a corresponding author.}
}

\markboth{IEEE Transactions on Pattern Analysis and Machine Intelligence}%
{Jeon \MakeLowercase{\textit{et al.}}: Measuring the Validity of Clustering Validation Datasets}

\IEEEpubid{}

\IEEEtitleabstractindextext{
\begin{abstract}
Clustering techniques are often validated using benchmark datasets where class labels are used as ground-truth clusters.
However, depending on the datasets, class labels may not align with the actual data clusters, and such misalignment hampers accurate validation.
Therefore, it is essential to evaluate and compare datasets regarding their cluster-label matching (CLM), i.e., how well their class labels match actual clusters.
Internal validation measures (IVMs), like Silhouette, can compare CLM over different labeling of the \textit{same dataset}, but are not designed to do so \textit{across different datasets}. 
We thus introduce \textit{Adjusted IVMs} as fast and reliable methods to evaluate and compare CLM across datasets.
We establish four axioms that require validation measures to be independent of data properties not related to cluster structure (e.g., dimensionality, dataset size).
Then, we develop standardized protocols to convert any IVM to satisfy these axioms, and use these protocols to adjust six widely used IVMs.
Quantitative experiments  (1) verify the necessity and effectiveness of our protocols and (2) show that adjusted IVMs outperform the competitors, including standard IVMs, in accurately evaluating CLM both \textit{within} and \textit{across datasets}.
We also show that the datasets can be filtered or improved using our method to form more reliable benchmarks for clustering validation.

\end{abstract}
}

\maketitle

\section{Introduction}

\label{sec:intro}

Cluster analysis \cite{jain88prenticehall} is an essential exploratory task for data scientists and practitioners in various application domains \cite{schaeffer07gc, lyi15bioinformatics, jeon25arxiv, caron18eccv, BonakalaDD2024, DarwishICWSM2020}. It commonly relies on unsupervised clustering techniques, that is, machine learning algorithms that partition data into subsets called groups or clusters. These algorithms maximize between-cluster separation and within-cluster compactness based on a given distance function \cite{liu10idcm}.

Clustering validation measures \cite{farber10multiclust,HassanCVIreview2024} or quality measures \cite{bendavid08nips} are used to evaluate clustering results. They are categorized as \textit{internal} measures and \textit{external} measures \cite{liu10idcm}. 
Internal validation measures (IVM) (e.g., Silhouette score~\cite{rousseeuw87jcam}), also known as relative measures \cite{jain88prenticehall}, give high scores to partitions in which data points with high or low similarities to each other are assigned to the same or different clusters, respectively. External validation measures (EVM) \cite{elhamifar13tpami, wu09kdd} (e.g., adjusted mutual information~\cite{vinh10jmlr}) quantify how well a clustering result matches an externally given ground truth partition.

\begin{figure}[t]
    \centering
    \includegraphics[width=\linewidth]{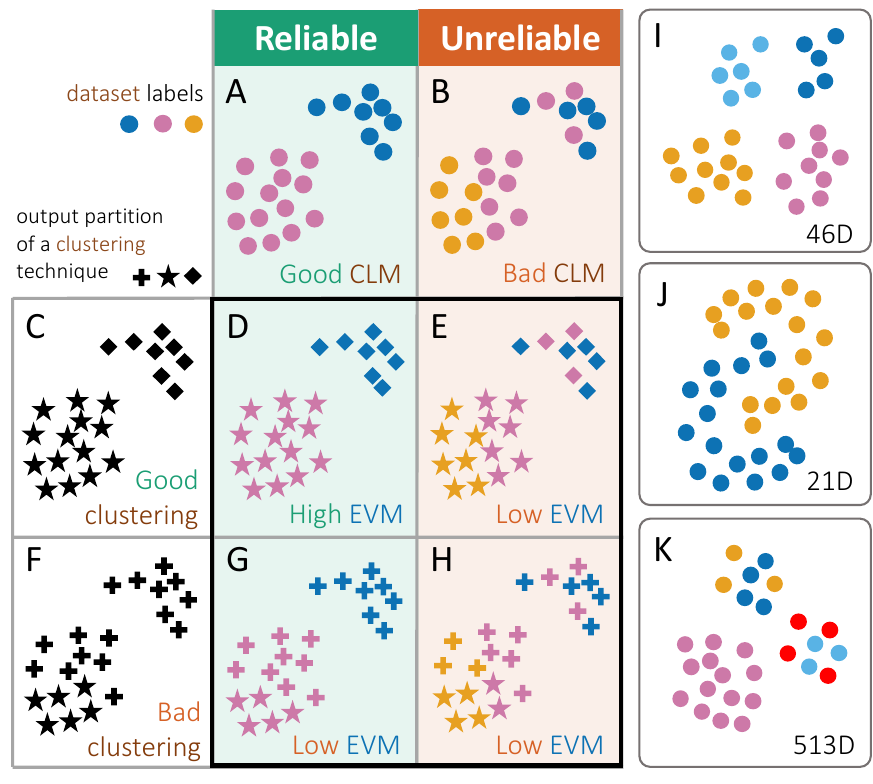}
    \caption{
    The illustration of how the degree of Cluster-Label Matching (\brown{CLM}) affects the reliability of external validation.
    An external validation measure (\blue{EVM}) evaluates how well the clustering results (markers' shape) match the ground truth partition, typically given by class labels (markers' hue). 
    The \brown{CLM} is \green{good} (A) if the partition formed by the \textit{class labels} (fill) matches well the \textit{clusters} formed by the point distribution (encoded by position). If \brown{CLM} is \green{good} (A), the \blue{EVM} between the dataset labels (A) and clustering results (C, F) gives a \green{\textit{reliable}} evaluation (D, G) of the \brown{\textit{clustering}} technique: \green{high}/\red{low} \blue{EVM} (D/G) aligns well with \green{good}/\red{bad} \brown{clustering} (C/F). But if \brown{CLM} is \red{bad} (B), the EVM is always \red{low} (E, H) and \red{\textit{unreliable}} to evaluate whether the \brown{clustering} is good (C) or bad (F). 
    We aim to evaluate and compare the \brown{CLM}  \textit{across datasets} (I,J,K) having diverse characteristics (e.g., dimension, size, data, and class distributions) to inform external validation by distinguishing valid benchmark datasets.
    }
    \label{fig:clm}
\end{figure}

Using the classes of a labeled dataset as a ground truth partition is a typical approach to conduct external validation \cite{farber10multiclust}. 
This approach promotes a clustering technique that precisely distinguishes labeled classes as separated clusters.
The underlying assumption  is that the classes of the dataset align well with the clusters \cite{aupetit14beliv, jeon23arxiv, farber10multiclust} (\autoref{fig:clm}A). We name it the \textit{Cluster-Label Matching (CLM)} assumption.

If the CLM of a labeled dataset is accurate, EVMs work as intended. EVM scores become low only when the clustering technique fails to capture the clusters represented by the classes (\autoref{fig:clm}G). That is, external validation is reliable only when the CLM assumption is valid.

\revise{However, datasets can have poor CLM due to labels split across clusters (\autoref{fig:clm}B magenta-colored points). For instance, images of buses and cars both assigned to the \texttt{Vehicle} class, may form distinct clusters in the image space. Conversely, data sets can also have poor CLM when they have multiple labels overlap within a single cluster (\autoref{fig:clm}B magenta- and yellow-colored points). This happens, for example,  when images of visually similar categories, such as leopards and cheetahs, cause \texttt{Leopard} and \texttt{Cheetah} classes to overlap in the image space.}

With such poor CLM datasets, EVMs become unreliable, producing low scores regardless of a clustering technique's capacity to capture clusters. In fact, a suboptimal clustering technique may receive a low EVM score because its incorrect cluster partition is unlikely to align well with the already inaccurate class label partition (\autoref{fig:clm}H). Conversely, even a well-performing clustering technique, despite correctly capturing the ``natural'' clusters, may also receive a low EVM score because these accurately identified clusters do not align well with class labels in bad-CLM datasets (\autoref{fig:clm}E).  
In essence, poor CLM undermines EVMs, making it difficult to differentiate between effective and ineffective clustering techniques.

It is thus crucial to evaluate CLM---the intrinsic validity of the ground truth labeled dataset---to compare clustering techniques on a reliable basis. 
Not checking CLM before executing external validation casts doubt on the results obtained. Moreover, such validation might lead to an erroneous conclusion when ranking and comparing clustering techniques (\autoref{sec:selection}).
Still, external validation is often conducted without considering the CLM of benchmark datasets \cite{khan21icecit, monath19kdd, masuyama22tpami, liu18tpami}. 
Our goal, therefore, is to evaluate the CLM of labeled datasets to distinguish credible benchmark datasets for clustering validation. We moreover aim to inform the community to use such datasets to conduct more reliable external validation. 

Yet, evaluating and comparing the CLM of datasets is a challenging problem. 
\revise{IVMs can be natural candidates for quantifying CLM} since well-clustered class partitions tend to naturally receive higher IVM scores. 
However, they are designed to compare different partitions of a \textit{single} dataset (\autoref{fig:clm}C \textit{vs.} \autoref{fig:clm}F), and not to compare partitions \textit{across different} datasets (\autoref{fig:clm}IJK). 
This limitation arises because IVMs depend not only on clustering quality but also on the dataset's size and dimensionality, and the distributions of data and class labels. Consequently, comparing IVM scores of different datasets is not reliable, making them improper for measuring and comparing CLM \textit{across} different datasets (\autoref{sec:bdsa}; \autoref{tab:between_rank}). 
\revise{This underscores the need for a more suitable measure of CLM to validate clustering benchmark datasets. }

To address this gap, we design \textit{adjusted} internal validation measures (\IVMa{}s) to assess and compare CLM \textit{across} datasets. \revise{Here are our key contributions:}

\begin{itemize}[leftmargin=1.2em]

\item \revise{ We first establish four \textit{across-dataset} axioms in \autoref{sec:axiom}, that form the theoretical basis for designing \IVMa{}s, complementing Ackerman and Ben-David's \textit{within-dataset} axioms \cite{bendavid08nips} already satisfied by standard IVMs. }

\item  \revise{ We propose adjustment protocols in \autoref{sec:tricks} for transforming an IVM into an \IVMa that satisfies each across-dataset axiom. 
\item In \autoref{sec:gener}, we use our protocols to adjust six of the most widely used IVMs \cite{liu10idcm} (e.g., Silhouette coefficient \cite{rousseeuw87jcam} and Calinski-Harabasz index \cite{calinski74cis}). The resulting \IVMa{}s are practical ways to quantify and compare CLM both within and across datasets.}

\item \revise{  In \autoref{sec:evaluation}, we conduct an ablation study to verify the validity and necessity of our new axioms and adjustment protocols. We also verify the effectiveness of \IVMa{}s in accurately evaluating and comparing CLM, both within and between datasets. Finally, a runtime analysis demonstrates the advantage of \IVMa{}s in terms of scalability.}

\item  \revise{Finally, we present two applications of these \IVMa{}s in \autoref{sec:applications} to demonstrate their benefits for the practitioners. We first validate the importance of evaluating the CLM over benchmark datasets by showing the instability of external validation when this step is overlooked. By doing so, we identify the top-CLM benchmark datasets that practitioners can use with higher confidence for evaluating clustering techniques. Second, we show that \IVMa{}s can also be used to correct bad-CLM benchmark datasets by searching for data subspaces that maximize \IVMa{}s.}

\end{itemize}

\section{Backgrounds and Related Work} 

\label{sec:relwork}

One common way to distinguish good and bad clustering techniques is to use EVMs. 
EVMs quantify how much the resulting clustering matches with a ground truth partition of a given dataset. For example, adjusted mutual information \cite{vinh10jmlr} measures the agreement of two assignments (clustering and ground truth partitions) in terms of information gain corrected for chance effects. 

The classes in labeled datasets have been used extensively as ground truth partitions for EVM \cite{farber10multiclust}. 
However, despite the potential risk of violating the CLM (\autoref{sec:intro}; \autoref{fig:clm}), no principled procedure has yet been proposed to evaluate the reliability of such a ground truth. Our research aims to fill this gap by proposing measures that can evaluate and compare CLM across datasets. A similar endeavor has been engaged in the supervised learning community to quantify datasets' difficulty for classification tasks \cite{pmlr-v162-ethayarajh22a}.

A natural approach is to use classification scores as a proxy for CLM \cite{abul03smc, rodriques18kbs}. This approach is based on the assumption that the classes of a labeled dataset with good classification scores would provide reliable ground truth for EVM. Still, a classifier can hardly distinguish between two ``adjacent'' classes forming a single cluster (\autoref{fig:clm}B orange and magenta points in the bottom left cluster, good class separation but bad CLM) and two ``separated'' classes forming distant clusters (\autoref{fig:clm}A blue and magenta clusters, good CLM). It also cannot distinguish different within-class structures, such as
a class forming a single cluster (\autoref{fig:clm}A blue class, good CLM) and one made of several distant clusters (\autoref{fig:clm}B magenta class, bad CLM). In addition, classifiers require expensive training time (\autoref{sec:runtime}).

A more direct approach is to examine how well the clustering techniques capture class labels, as well-separated classes will be easily captured by the techniques  (\autoref{fig:clm}D). However, this approach is also computationally expensive (\autoref{sec:runtime}).
Moreover, it is not based on principled axioms independent of any clustering technique, so it is likely to be biased with respect to a certain type of cluster. Still, we can approximate ground truth CLM by running multiple and diverse clustering techniques \cite{vega11jprai} and aggregating their EVM scores.  For lack of a better option, we use this ensemble approach to obtain an approximate ground truth in our experiments to validate our axiom-based solution.

In contrast, IVMs are inexpensive to compute (\autoref{sec:runtime}). They also examine cluster structure in more detail, relying on two criteria, namely \textit{compactness} (i.e., the pairwise proximity of data points within a cluster) and \textit{separability} (i.e., the degree to which clusters lie apart from one another) \cite{liu10idcm, tan05idm}. For example, in \autoref{fig:clm}, an IVM would give a higher score to clustering partition C than to F. 
Moreover, following the axiomatization of clustering by Kleinberg \cite{kleinberg02nips}, Ackerman and Ben-David \cite{bendavid08nips} proposed four \textit{within-dataset} axioms that give a common ground to IVMs: scale invariance, consistency, richness, and isomorphism invariance (\autoref{sec:withinaxiom}). These axioms set the general requirements that a function should satisfy to work properly as an IVM.

Nevertheless, IVMs were originally designed to compare and rank different partitions of the \textit{same} dataset (\autoref{fig:clm}A-H). 
Therefore, IVMs are not only affected by cluster structure but also dependent on the characteristics of the datasets, such as the number of points, classes, and dimensions (\autoref{fig:clm}I-K), which means that they are cannot properly compare CLM \textit{across} different datasets. 
Here, we propose four additional axioms that IVMs should satisfy to compare CLM across datasets and derive new \textit{adjusted} IVMs satisfying the axioms (i.e., \IVMa{}s). 
\IVMa{}s play the role of a proxy of CLM, evaluating the intrinsic validity of a benchmark dataset for external clustering validation.
\section{New Axioms for Adjusted IVM}

\label{sec:axiom}

We propose new \textit{across-dataset} axioms that adjusted IVMs (i.e., \IVMa{}s) should satisfy to properly evaluate and compare CLM across datasets, complementing the within-dataset ones of Ackerman and Ben-David \cite{bendavid08nips}.

\subsection{\textit{Within-Dataset} Axioms}

\label{sec:withinaxiom}

Ackerman and Ben-David (\AB) introduced \textit{within-dataset} axioms \cite{bendavid08nips} that specify the requirements for IVM to properly evaluate clustering partitions (Appendix A): \textbf{W1: Scale Invariance} requires measures to be invariant to distance scaling; \textbf{W2: Consistency} is satisfied by a measure that increases when within-cluster compactness or between-cluster separability increases; \textbf{W3: Richness} requires measures possible to give any fixed cluster partition the best score over the domain by only modifying the distance function; and \textbf{W4: Isomorphism Invariance} ensures that an IVM does not depend on the external identity of points (e.g., class labels).

\subsection{\textit{Across-Dataset} Axioms}

\label{sec:betweenaxiom}

Within-dataset axioms do not consider the case of comparing scores across datasets; rather, they assume that the dataset is invariant. 
We propose four additional \textit{across-dataset} axioms that a function should satisfy to fairly compare cluster partitions across datasets.

\paragraph{Notations}
\revise{We begin by defining four fundamental building blocks of our axioms, using notation identical to that of \AB:
\vspace{3pt}
\begin{itemize}[leftmargin=1.2em]
    \item A \textbf{finite domain set} $X \subset \mathcal{D}$ of dimension $\Delta_X$, where $\mathcal{D}$ denotes the data space.
    \item A \textbf{clustering partition} of $X$ as $C =\{C_1, C_2, \cdots, C_{|C|}\}$, where  $\forall i\neq j, C_i \cap C_j = \emptyset$ and $\cup_{i=1}^{|C|} C_i = X$.
    \item A \textbf{distance function} $\delta : \mathcal{D} \times \mathcal{D} \rightarrow \mathbb{R}$, satisfying $\delta(x, y) \geq 0$, $\delta(x, y) = \delta(y, x)$ and $\delta(x, y) = 0$ if $x=y$ $\forall x, y \in \mathcal{D}$. We do not require the triangle inequality.
    \item A \textbf{measure} $f$ as a function that takes $C, X, \delta$ as input and returns a real number. Higher $f$ implies a better clustering. 
\end{itemize}
\vspace{3pt}
We extend these notations with:
\vspace{3pt}
\begin{itemize}[leftmargin=1.2em]
    \item $\overline{X'}$ the \textbf{centroid} of $X'$, where $X' \subset X$. 
    \item $\underline{W}_\alpha$ a \textbf{random subsample}  of the set $W$ ($\underline{W}_\alpha\overset{D}{=}W$) such that $|\underline{W}_\alpha|/|W|=\alpha$, and the corresponding clustering partition is noted $\underline{C}_\alpha=\{\underline{C_i}_\alpha\}_{i=1, \dots, |C|}.$
\end{itemize}
}

\paragraph{Goals and factors at play}
\AB's within-dataset axioms are based on the assumption that the measures that satisfy these axioms properly evaluate the quality of a clustering partition.
However, the axioms do not consider that the measures $f$ could operate on varying $C$, $\delta$, and $X$. For example, isomorphism invariance (W4) assumes fixed $X$ and $\delta$; consistency (W2) and richness (W3) define how functions $f$ should react to the change of $\delta$, but do not consider how $\delta$ changes in real terms, affected by various aspects of $X$ (e.g., dimensionality); scale invariance (W1) considers such variations, but only in terms of the global scaling. Thus, the satisfaction of \AB's axioms is a way to ensure IVMs focus on measuring clustering quality within a single dataset but not across datasets.

In contrast, \IVMa{}s shall operate on varying $C$, $\delta$, and $X$. Thus, 
several aspects of the varying datasets now come into play, and their influence on \IVMa shall be minimized. The sample size $|X|$ is one of them (\textbf{Axiom A1}), and the dimension $\Delta_X$ of the data is another one (\textbf{Axiom A2}). 
Moreover, what matters is the \textit{matching} between natural clusters and data labels more than the number of clusters or labels; therefore, 
we shall reduce the influence of the number of labels $|C|$ (\textbf{Axiom A3}). Lastly, we need to align \IVMa to a comparable range of values (\textbf{Axiom A4}) across datasets, in essence capturing all remaining hard-to-control factors unrelated to clustering quality. We now explain the new axioms in detail:

\subsubsection*{\revise{Axiom A1: Data-Cardinality Invariance}}
Invariance of the sample size $|X|$ is ensured if subsampling all clusters in the same proportion does not affect the \IVMa score. This leads to the first axiom:

\paragraph{\textbf{A1 -- Data-Cardinality Invariance}} \textit{A measure $f$ satisfies data-cardinality invariance if
$\forall X,\forall \delta$ and $\forall C$ over $(X,\delta)$, $f(C,X,\delta) = f(\underline{C}_\alpha,X_\alpha,\delta)$ with $X_\alpha=\cup_{i=1}^{|C|} \underline{C_i}_\alpha$  $\forall \alpha\in ]0,1]$.}

\subsubsection*{\revise{Axiom A2: Shift Invariance}} 
We shall consider that data dimension $\Delta_X$ varies across datasets. An important aspect of the dimension called the concentration of distance phenomenon, which is related to the curse of dimensionality, affects the distance measures involved in \IVMa. As the dimension grows, the variance of the pairwise distance for any data tends to be constant, while its mean value increases with the dimension \cite{beyer99icdt, francois07tkde, lee11iccs}. Therefore, in high-dimensional spaces, $\delta$ will act as a constant function for any data $X$, and thus an \IVMa $f$ will generate similar scores for all datasets. 
To mitigate this phenomenon, and as a way to reduce the influence of the dimension, we require that the measure $f$ be shift invariant \cite{lee11iccs, lee14cidm} so that the shift of the distances (i.e., growth of the mean) can be canceled out.

\paragraph{\textbf{A2 -- Shift Invariance}} \textit{A measure $f$ satisfies the shift invariance if $\forall X, \forall \delta$, and  $\forall C$ over $(X, \delta)$, $f(C, X, \delta) = f(C, X, \delta+ \beta)$ $ \forall \beta > 0$, where $\delta + \beta$ is a distance function satisfying $(\delta + \beta)(x,y) = \delta(x,y) + \beta$, $\forall x, y \in X$.}

\subsubsection*{\revise{Axiom A3: Class-Cardinality Invariance}} The number of classes should not affect an \IVMa; for example, two well-clustered classes should get an \IVMa score similar to 10 well-clustered classes. \AB proposed that the minimum, maximum, and average class-pairwise aggregations of IVMs form yet other valid IVMs. We follow this principle as an axiom for \IVMa.

\sloppy
\paragraph{\textbf{A3 -- Class-Cardinality Invariance}} 
\textit{A measure $f$ satisfies class-cardinality invariance if $\forall X,  \forall \delta$ and $\forall C$ over $(X, \delta)$, $f(C, X, \delta) = \textrm{agg}_{S \subseteq C, |S| = 2} f'(S, X, \delta)$ \textit{where function}  $\textrm{agg}_S\in\{\textrm{avg}_S,\min_S,\max_S\}$ and $f'$ is an IVM.}

\fussy

\subsubsection*{\revise{Axiom A4: Range Invariance}} Lastly, we need to ensure that an \IVMa  takes a common range of values across datasets. In detail, we want their minimum and maximum values to correspond to the datasets with the worst and the best CLM, respectively, and that these extrema are aligned across datasets (we set them arbitrarily to 0 and 1), as follows:

\sloppy
\paragraph{\textbf{A4 -- Range Invariance}} \textit{A measure $f$ satisfies range invariance if $\forall X, \forall \delta$, and $\forall C$ over $(X,\delta)$, $\min_{C} f(C,X,\delta)=0$ and $\max_{C} f(C,X,\delta)=1$.}

\fussy

\section{Generalization Protocols}

\label{sec:tricks}

We introduce four technical protocols (T1-T4), designed to generate \IVMa{}s that satisfy the corresponding axioms  A1-A4, respectively.

\subsubsection*{T1: Approaching Data-Cardinality Invariance (A1)}

We cannot guarantee the invariance of a measure for any subsampling of the data (e.g., very small sample size). However, we can obtain robustness to random subsampling if we use consistent estimators of population statistics \cite{VaartAsymptoticStat1998} as building blocks of the measure. For example, we can use the mean, the median, or the standard deviation of the points within a class or the whole dataset or quantities derived from them, such as the average distance between all points of two classes.

\subsubsection*{T2: Achieving Shift Invariance (A2)}

\noindent
\textbf{T2-a,b: Exponential protocol.} Considering a vector of distances $u=(u_1\dots u_n)$, we can define a shift-invariant function by using a ratio of exponential functions: 
\begin{equation}
    g_j(u)=\frac{e^{u_j}}{\sum_k e^{u_k}}.
\end{equation}
We observe that $\forall \beta \in \mathbb{R}$,
\begin{equation}
g_j(u+\beta) = \frac{e^{u_j+\beta}}{\sum_k e^{u_k+\beta}}=\frac{e^{u_j}}{\sum_k e^{u_k}}\frac{e^{\beta}}{e^{\beta}}=g_j(u), 
\end{equation}
hence $g_j$ is shift invariant. Thus, the measure $f$ is shift invariant if it consists of ratios of the exponential distances (\textbf{T2-a}). Note that this protocol is at the core of the $t$-SNE loss function \cite{lee11iccs}.
If a building block is a sum or average of distances, the exponential should be applied to the average of distances rather than individuals (\textbf{T2-b}), as the shift occurs to the average distances \cite{francois07tkde}.

\paragraph{T2-c: Equalizing shifting}
The exponential protocol can be safely applied only if the measure incorporates the distance between data points within $X$ (Type-1 distance).
We do not know, in general, how the shift of type-1 distances affects the distances between data points and their centroid (Type-2),
nor do we know how the shift affects the distance between two centroids (Type-3),
even though they are common building blocks in IVM \cite{liu10idcm}. Fortunately, if $\delta$ is the square of Euclidean distances (i.e., $\delta =d^2$ where $d(x,y)$ denotes the Euclidean distance between points $x$ and $y$), we can prove that the shift of type-1 distances by $\beta$ results in the shift of type-2 distances by $\beta/2$, and in no shift of type-3 distances, which is stated by the following theorems (proof in Appendix D.1).

\paragraph{Theorem 1 (Type-2 Shift)} \textit{$\forall X' \subset X$,  $\forall \beta > 0$, and for any Euclidean distance functions $d_L$ and $d_H$ satisfying $d_H^2 = d_L^2 + \beta$, 
    $\sum_{x \in X'}d_H^2(x, c) = \sum_{x \in X'}d^2_L(x,c) + \beta/2$, where $c = \overline{X'}$.}
    
\paragraph{Theorem 2 (Type-3 Shift).} \textit{$\forall X', X'' \subset X$, $\forall \beta > 0$, and for any Euclidean distance functions $d_L$ and $d_H$ satisfying $d_H^2 = d_L^2 + \beta$, $d^2_H(c', c'') = d^2_L(c', c'')$, where $c' = \overline{X'}$ and $c''= \overline{X''}$.}
\vspace{4pt}

\noindent
Therefore, if an IVM consists of different types of distances, we should use $\delta = d^2$ and apply the exponential protocol with the same type of distances for both its numerator and denominator (\textbf{T2-c}).

\paragraph{T2-d: Recovering Scale Invariance}
After applying the exponential protocol, $g_j$ is no more scale-invariant:
\revise{
\begin{equation}
    \forall \lambda\in \mathbb{R},g_j(\lambda u) = \frac{e^{\lambda u_j}}{\sum_k e^{\lambda u_k}}\neq g_j(u),
\end{equation}
}
and so it will not satisfy axiom W1. We can recover scale-invariance by normalizing each distance $u_i$ by a term that scales with all of the $u_k$ together, such as their standard deviation, $\sigma(u)$. Now, 
\revise{
\begin{equation}
    g_j(\lambda u/\sigma(\lambda u))=g_j(\lambda u/\lambda \sigma(u))=g_j(u/\sigma(u)),
\end{equation}
}
is both shift and scale invariant (\textbf{T2-d}).

\subsubsection*{T3: Achieving Class-Cardinality Invariance (A3)}

\sloppy
Class-cardinality invariance can be achieved by following the definition of Axiom A3; thas is, by defining the global measure as the aggregation of class-pairwise local measures, $f_c(C, X, d) = \textrm{agg}_{S \subseteq C, |S| = 2} f(S, X, \delta)$, where $\textrm{agg} = \{\textrm{avg}, \min, \max\}$.

\fussy

\subsubsection*{T4: Achieving Range Invariance (A4)}

\label{sec:trick4}

\noindent
\textbf{T4-a,b: Scaling.} 
A common approach to get a unit range for $f$ is to use min-max scaling $f_u=(f-f_{\min})/(f_{\max}-f_{\min})$.
However, determining the minimum and maximum values of $f$ for any data $X$ is nontrivial. Theoretical extrema are usually computed for edge cases far from realistic $X$ and $C$. Wu et al. \cite{wu09kdd}  proposed estimating the worst score over a given dataset $X$ by the expectation $\hat{f}_{\min}=E_{\pi}(f(C^{\pi},X,\delta))$ of $f$ computed over random partitions $C^{\pi}$ of $(X,\delta)$ preserving class proportions $|C^{\pi}_i|=|C_i|{\forall i}$ (\textbf{T4-a}), which are arguably the worst possible clustering partitions of $X$. In contrast, it is hard to estimate the maximum achievable score over $X$, as this is the very objective of clustering techniques. If the theoretical maximum $f_{\max}$ is known and finite, we use it by default; otherwise, if $f_{\max}\rightarrow+\infty$ then the scaled measure $f_u \rightarrow 0, \forall f$. We propose to use a logistic function $f'= 1 / (1 + e^{-k\cdot f})$ (\textbf{T4-b}) before applying the normalization so $f'_{\max}=1$ and $f'_{\min}=\hat{f'}_{\min}$.

\paragraph{T4-c: Calibrating logistic growth rate \textit{k}} 
We can arbitrarily make a logistic function to pull or push all scores toward the minimum or maximum value by tuning the growth rate $k$. 
We thus propose calibrating $k$ with datasets with ground truth CLM scores. 
 Assume a set of labeled datasets $\mathcal{X} = \{X^{1}, \cdots, X^{n}\}$ with class labels
 $\mathcal{C} = \{C^{1}, \cdots, C^{n}\}$ and the corresponding ground truth CLM scores
 $\mathcal{S} = \{s^{1}, \cdots, s^{n}\}$ where $\min \mathcal{S} =0$ (worst) and 
 $\max \mathcal{S} = 1$ (best). Here, we can optimize $k$ to make
 $\mathcal{S}' = \{s'^{1}, \cdots, s'^{n}\}$ best matches with $S$, where 
 $s'^{i} = f(C^{i}, X^{i}, \delta)$. In practice, we use Bayesian optimization \cite{snoek12neurips} targeting the $R^2$ score. 

 We propose using human-driven separability scores as a proxy for CLM, building upon available human-labeled datasets acquired from a user study \cite{abbas19cgf} and used in several works on visual perception of cluster patterns \cite{AupetitVIS19},\cite{Abbas2024},\cite{jeon24tvcg}. Each dataset consists of a pair of Gaussian clusters (classes) with diverse hyperparameters (e.g., covariance, position), graphically represented as a monochrome scatterplot. The perceived separability score of each pair of clusters was obtained by aggregating the judgments of 34 participants of whether they could see one or more than one cluster in these plots. The separability score of each dataset is defined as the proportion of participants who detected more than one cluster. 
We used these datasets $\mathcal{X}$ and separability scores $\mathcal{S}$ for calibration because they are not biased by a certain clustering technique or validation measure; they are based on human perception following a recent research trend in clustering \cite{AupetitVIS19},\cite{BlasilliVisualCQM2024}, and the probabilistic scores naturally range from 0 to 1.
 However, as the scores are not uniformly distributed, we bin them and weigh each dataset in proportion to the inverse size of the bin they belong to (see Appendix B for details).

\section{Adjusting IVM into \IVMa}

\label{sec:gener}

We use the proposed protocols (\autoref{sec:tricks}) to adjust six baseline IVMs: Calinski-Harabasz index ($CH$) \cite{calinski74cis}, Dunn index ($DI$) \cite{dunn74joc}, I index ($II$) \cite{maulik02tpami}, Xie-Beni index ($XB$) \cite{xie91tpami}, Davies-Bouldin index ($DB$) \cite{davies79tpami}, and Silhouette coefficient ($SC$) \cite{rousseeuw87jcam}, into \IVMa{}s that satisfy both within- and across-dataset axioms. 
We pick the IVMs from the list of the survey done by Liu et al. 
\cite{liu10idcm}. 
We select every IVM except the ones optimized based on the elbow rule (e.g., modified Hubert $\Gamma$ statistic \cite{hubert85classification}) and those that require several clustering results (e.g., S\_Dbw index \cite{halkidi01icdm}). 
Our choice covers the most widely used IVMs that have clear variety in the way of examining cluster structure.

Here, we explain the adjustment of $CH$. We selected $CH$ because it does not satisfy any of the across-dataset axioms, so that we can demonstrate the application of all protocols (\autoref{sec:tricks}). The adjusted $CH$ ($CH_A$) also turned out to be the best \IVMa in our evaluations (\autoref{sec:evaluation}).

\subsection{Adjusting the Calinski-Harabasz Index}

$CH$ \cite{calinski74cis} is defined as:
\begin{equation}
 CH(C, X, d^2) = \displaystyle\frac{|X| - |C|}{|C| - 1}\cdot\frac{\sum_{i=1}^{|C|}|C_i|d^2(c_i, c)}{\sum_{i=1}^{|C|} \sum_{x \in C_i} d^2(x, c_i)}, 
\end{equation}
where $c_i=\overline{C_i}$ and $c = \overline{X}$. A higher value implies a better CLM. The denominator and numerator measure compactness and separability, respectively. The adjustment procedure is as follows:

\paragraph{Applying T1 (Data-cardinality invariance)}
Both the denominator and numerator of $CH$ are already robust estimators of population statistics (T1). However, as the term $(|X|-|C|)$ makes the score grow proportional to the size of the datasets, we remove the term to eliminate the influence of data-cardinality, resulting in: 
\begin{equation}
 CH_{1}(C, X, d^2) = \displaystyle\frac{\sum_{i=1}^{|C|}|C_i|d^2(c_i, c)}{(|C| - 1)\sum_{i=1}^{|C|} \sum_{x \in C_i} d^2(x, c_i)}.
\end{equation}

\paragraph{Applying T2 (Shift invariance)}
$CH_1$'s numerator and denominator consists of \mbox{type-3}  and type-2 distances, respectively. To equalize the shift before applying exponential (T2-c), we add the sum of the squared distances of the data points to their centroid as a factor to the numerator, which does not affect separability or compactness. This leads to:\
\begin{equation}
\displaystyle
CH_{2}(C, X, d^2) = \displaystyle\frac{\displaystyle\sum_{x \in X} d^2(x, c)}{\displaystyle\sum_{i=1}^{|C|} \displaystyle\sum_{x \in C_i} d^2(x, c_i)} \cdot \frac{\displaystyle\sum_{i=1}^{|C|}|C_i|d^2(c_i, c)}{|C| - 1}.
\end{equation}
As the left term is a fraction of the sum of type-2 distances, we get shift invariance by dividing both the numerator and the denominator by $|X|$ (i.e., the sum becomes an average; T2-b), then by applying the exponential normalized by the standard deviation $\sigma_{d^2}$ of the square distances of the data points to their centroid (T2-a); i.e., $\sigma_{d^2} = \text{std}(\{d^2(x, c) | x \in X\})$.
The right term does not need an exponential protocol as type-3 distances do not shift as the dimension grows. We still divide the term with $|X|$ and $\sigma_{d^2}$ to ensure data-cardinality and scale invariance, respectively. This leads to:
\begin{equation}
\begin{split}
CH_{3}(C, X, d^2) &= \frac{e^{\sum_{x \in X} \frac{d^2(x, c)}{\sigma_{d^2} \cdot |X|}}}{e^{\sum_{i=1}^{|C|} \frac{\sum_{x \in C_i} d^2(x, c_i)}{\sigma_{d^2} \cdot|X|}}} \cdot \\ &\quad \times \frac{\sum_{i=1}^{|C|}|C_i|d^2(c_i, c)}{\sigma_{d^2} \cdot|X| \cdot (|C| - 1)}.
\end{split}
\end{equation}



\paragraph{Applying T4 (Range invariance)}
We apply min-max scaling to make the measure range invariant. 

As $\max(CH_3) \rightarrow +\infty$, we transform it through a logistic function (T4-b), resulting in:
\revise{
\begin{equation}
    CH_4 = 1 / (1 +e^{-k \cdot CH_3}), \text{ }\therefore CH_{4\max} \rightarrow 1.
\end{equation}}
We then estimate the worst score $CH_{4\min}$ (T4-a) as the average $CH_4$ score computed over $T$ Monte-Carlo simulations with random clustering partitions $C^{\pi}$: 
\revise{
\begin{equation} \label{eq:montecarlo}
    CH_{4\min}=\frac{1}{T}\sum_{t=1}^T CH_4(C^{\pi_t}, X, d^2),
\end{equation}}.

We then get:
\revise{
\begin{equation}
    CH_5= (CH_4 - CH_{4\min}) / (CH_{4\max} - CH_{4\min}),
\end{equation}}
where we set the logistic growth rate $k$ by calibrating the $CH_5$ scores ($\mathcal{S'}$) with human-judgment scores ($\mathcal{S}$) (T4-c).

\paragraph{Applying T3 (Class-cardinality invariance)}
Lastly, we make our measure satisfy class-cardinality invariance (Axiom A3) by averaging class-pairwise scores (T3), which finally determines the adjusted Calinski-Harabasz index:
\begin{equation}
CH_A(C, X, d^2) = \frac{1}{{|C| \choose 2}}\sum_{S \subseteq C, |S| = 2} CH_5(S, X, d^2). 
\end{equation}
Unlike $CH$, which misses all across-dataset axioms, $CH_A$ satisfies all of them (Refer to Appendix D for the proofs).

\paragraph{Removing Monte-Carlo simulations}
We can reduce the computing time of $CH_A$ by removing Monte-Carlo Simulations for estimating $CH_{4\min}$. Indeed, as randomly permuting class labels make all $C_i \in C$ satisfy $C_i \overset{D}{=} X$, we can assume $c \simeq c_i$ $\forall c_i$. Therefore, $CH_3(C^{\pi}, X, d^2) \simeq 0$ as it contains $d^2(c_i, c) \simeq 0$ in the second term, which leads to:
\revise{
\begin{equation} 
\begin{split}
    CH_{4\min} &= E_\pi(CH_4(C^{\pi}, X, d^2)) \\ &= E_\pi(1/2) = 1/2.
\end{split}
\end{equation}
}
This approximation also makes $CH_A$ deterministic.

\paragraph{Computational complexity}
\revise{
\autoref{tab:comp} presents the time complexities of the IVMs. 
Since the complexity of $CH_A$ is linear with respect to all parameters, where the only additional parameter compared to $CH$ is $|C|$, the measure scales efficiently to datasets with large sizes and high dimensionality (\autoref{sec:runtime}). 
}

\begin{table}[t]
\centering
\caption{\revise{Time complexity of $CH$, $CH_A$ and their variants generated by applying our protocols (\autoref{sec:gener}). The simulation refers to $T$ runs of Monte-Carlo experiments used to compute $CH_{4\min}$ \autoref{eq:montecarlo}). IVMs in the first two rows do not depend on Monte-Carlo simulations. }}
\label{tab:comp}
\begin{tabular}{lll}
\toprule
\textbf{IVMs} & \textbf{w/ simulation} & \textbf{w/o simulation} \\
\midrule
$CH, CH_1, CH_2, CH_3, CH_4$ & \multicolumn{2}{c}{$O(|X|\Delta_X)$} \\
$CH_{4\max}$ & \multicolumn{2}{c}{$O(1)$} \\
$CH_{4\min}$ & $O(T|X|\Delta_X)$ & $O(|X|\Delta_X)$ \\
$CH_5$ & $O(T|X|\Delta_X)$ & $O(|X|\Delta_X)$ \\
$CH_A$ & $O(T|X|\Delta_X\,|C|)$ & $O(|X|\Delta_X \,|C|)$ \\
\bottomrule
\end{tabular}
\end{table}

\begin{figure*}
    \centering
    \includegraphics[width=\textwidth]{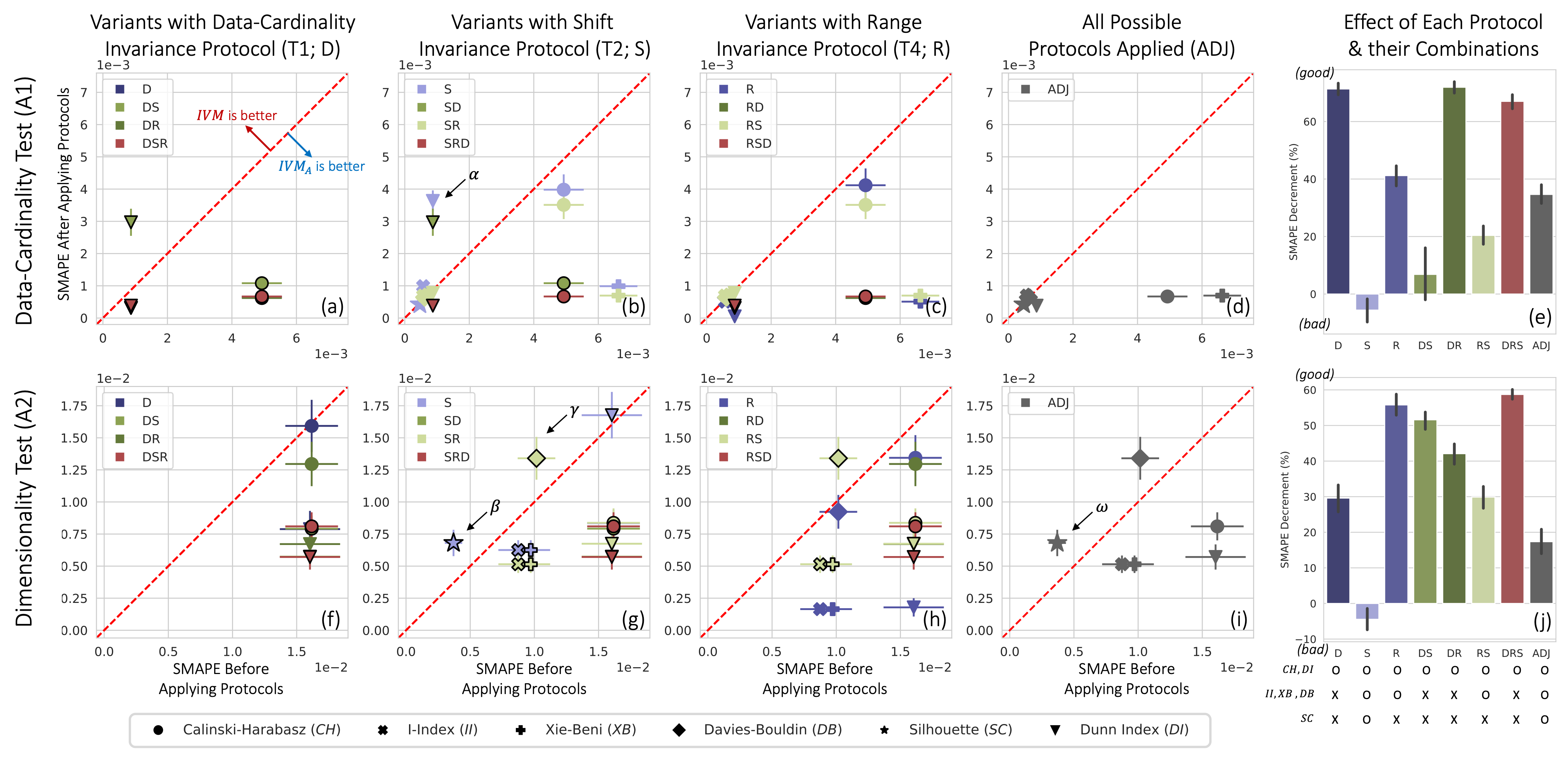}
    \caption{Results of the ablation study (\autoref{sec:validity}). Top: data-cardinality test, bottom: dimensionality test.  
    Scatterplots (a-d, f-i) show how the SMAPE varies between IVMs and their variants in average, where each dot corresponds to a single IVM variant $Z_v$. The shape and the color of the dots correspond to $Z$ and $v$, respectively, and $x$ and $y$ coordinates represent the error made by $Z$ (before applying the protocols) and $Z_v$ (after applying the protocols), respectively. 
    If a dot is located in the lower-right half of the scatterplot, it means that the protocol-based variant $Z_v$ produces less error than $Z$.  
    Note that $II$ and $XB$ are horizontally aligned as their adjusted versions are identical. 
    The first three columns show the effect of each protocol and its variants, and the fourth column shows the variant in which all possible protocols are applied ($Z_v = Z_A$). Bar plots (e, j) show the error reduction rate made by the protocols and their combinations averaged over all the IVMs; the higher the bar, the better. Notice that ADJ shows lower error reduction compared to other combinations because not all IVMs benefit from the entire set of protocols. 
The table below the bar plots indicates what combinations of protocols are applied (O) to each IVM or not (X).
    }
    \label{fig:ablation}
    \vspace{-2mm}
\end{figure*}

\subsection{Adjusting the Remaining IVMs}

\label{sec:remaining}

The adjustment processes of the remaining IVMs are not significantly different from those of $CH$.
As $DI$ misses all across-dataset axioms, it goes through all protocols like $CH$. $II$, $XB$, and $DB$ require the shift (T2), range (T4), and class-cardinality (T3) invariance protocols as they are already data-cardinality invariant (A1). After passing through these protocols, $II$ and $XB$ become identical (i.e., $II_A = XB_A$). For $SC$, only shift and class-cardinality invariance protocols are required, as data-cardinality and range invariance (A4) are already satisfied. We thus obtain five adjusted IVMs: $CH_A$, $DI_A$, $\{II, XB\}_A$, $DB_A$, and $SC_A$. Please refer to Appendix C for detailed adjustment processes.

\section{Evaluation}

\label{sec:evaluation}

\revise{
We conduct four experiments to evaluate our protocols (\autoref{sec:tricks}) and \IVMa{}s. The first experiment evaluates the effectiveness of each individual protocol by an ablation study (\autoref{sec:validity}). The second and third experiments investigate how well \IVMa correlates with the ground truth CLM across (\autoref{sec:bdsa}) and within (\autoref{sec:wdca}) datasets, comparing their performance to competitors like standard IVMs and supervised classifiers (\autoref{sec:relwork}).
Finally, we investigate the runtime of \IVMa{}s and competitors in quantifying the CLM of datasets (\autoref{sec:runtime}). 
}

\subsection{Ablation Study on the Protocols}
\label{sec:validity}

\noindent
\textbf{Objectives and design.}
We want to evaluate our protocols in making \IVMa{}s to satisfy the new axioms. 
For datasets, we consider the ones that we used to fit the logistic growth rate $k$ (\autoref{sec:trick4}, T4-c), which consists of two bivariate Gaussian clusters (as classes) with various levels of CLM \cite{abbas19cgf}, to which we add noisy dimensions. 
For each \IVMa $Z_A$, we consider its variants $Z_v$ made by switching \textit{on} and \textit{off} the protocols ($Z_v \in \{Z_P | P \in 2^{\mathcal{T}_Z}\}$ where $\mathcal{T}_Z$ is the set of protocols switched for $Z$). 
In the case of $CH$ and $DI$, we make eight variants by switching data-cardinality (T1; D), shift invariance (T2; S), and range invariance (T4; R) protocols ($\mathcal{T}_{CH} = \mathcal{T}_{DI} = \{T1, T2, T4\}=\{D,S,R\}$), 
and for $II$, $XB$, $DB$, we switch T2 and T4 ($\mathcal{T}_{II} = \mathcal{T}_{XB} = \mathcal{T}_{DB} = \{T2, T4\}= \{S, R\}$), resulting in four variants. 
For $SC$, only T2 is turned \textit{on} and \textit{off}, which leads to two variants ($\mathcal{T}_{SC} = \{T2\} = \{S\}$).
The effect of the class-cardinality invariance protocol (T3) is not evaluated because the ground truth synthetic datasets \cite{abbas19cgf} contain only two classes. 
We control the cardinality (A1) and dimension (A2) of the datasets to evaluate how sensitive the variants are to variations of these conditions (the lower, the better). We do not control class-cardinality (A3) as the number of classes (which is 2) is imposed by the available data. Range invariance (A4) is not controlled either, as it is imposed by the min-max protocol (T4) and is not a characteristic of the datasets.

\paragraph{Datasets} We use 1,000 base datasets $\{X_1\dots X_{1000}\}$, each consisting of $|X|=1,000$ points sampled from two Gaussian clusters ($|C|=2$) within the 2D space and augmented with $98$ noisy dimensions ($\Delta=100$). We control the eight independent parameters (ip) of the Gaussians: two covariance matrices (3 ip each), class proportions (1 ip), and the distance between Gaussian means (1 ip), following previous studies \cite{abbas19cgf, AupetitVIS19}. New to this work, we add Gaussian noise along the supplementary dimensions, to each cluster-generated data, with mean 0 and variance equal to the minimum span of that cluster's covariance. We generate any dataset $X_{i,t}$ by specifying a triplet $(X_i,N_t,\Delta_t)$ with $X_i$ a base dataset, $N_t$ the number of data randomly sampled from $X_i$ preserving cluster proportions, and $\Delta_t$ its dimension where the first two dimensions always correspond to the 2D cluster space.
\textbf{Sensitivity to data-cardinality (A1):} (\autoref{fig:ablation} top) 
For each of the 1,000 base data $X_i$, we generated $11$ datasets $X_{i,t}=(X_i,N_t,\Delta_t)_{i\in[1\dots1000],t\in[0\dots10]}$, with the controlled data cardinality set to $N_t=50t+500$ and $\Delta_t$ drawn uniformly at random from $[2,\dots ,100]$.
\textbf{Sensitivity to dimensionality (A2):} (\autoref{fig:ablation} bottom)
For each of the $1000$ base data $X_i$, we generated $11$ datasets $X_{i,t}=(X_i,N_t,\Delta_t)_{i\in[1\dots1000],t\in[0\dots10]}$, with $N_t$ drawn uniformly at random from $[500,\dots,1000]$ and the controlled dimension set to $\Delta_0=2$ or $\Delta_t=10t, \forall t>0$.

\paragraph{Measurements} \sloppy
\revise{
We quantify the extent to which each variant of a score changes due to the feature alteration of base datasets.}
Formally, for each variant $Z_v$, we evaluate the matching between a pair ($a,b)$ of values of the controlled factor $t$ (\textit{e.g.} ($\Delta_a,\Delta_b)=(10,30)$) across all the 1,000 base datasets using: 
\revise{
\begin{equation}
\begin{split}
S_{k \in \{1,\dots,1000\}} \bigl(
    &w_k\bigl(Z_v(C_k, X_{k,a}, d) - M\bigr),\\
    &w_k\bigl(Z_v(C_k, X_{k,b}, d) - M\bigr)
\bigr).
\end{split}
\end{equation}
}
where S is the symmetric mean absolute percentage error (SMAPE) \cite{tofallis15jors}, which is defined as
\revise{
\begin{equation}
    {\displaystyle {S_{k\in K} (F_k,G_k)}=\frac{\sum _{k\in K} |F_{k}-G_{k}|}{\sum _{k\in K}(|F_{k}|+|G_{k}|)}}
\end{equation}
}
(0 best, 1 worst), 
and \revise{$M$ is the overall minimum empirical score across all base datasets:}
\revise{
\begin{equation}
\begin{split}
    M = \min\bigl(&\min_{k\in[1\dots 1000]}Z_v(C_{k},X_{k,a},d), \\&\min_{k\in[1\dots 1000]} Z_v(C_{k},X_{k,b},d)\bigr).
\end{split}
\end{equation}
}
Note that $C_k$ is the class label of $X_k$ and $w_k$ is the weight of $X_{k}$ to mitigate the distribution imbalance, which is also used to adjust the growth rate in the range invariance protocol (T4-c; Appendix B).
We adapt SMAPE to compare measures with different ranges equally and align scores to 0 by subtracting $M$, as SMAPE 
can be over-forecasted by shifting $F_k$ and $G_k$ to the positive side \cite{putz21re}.

\fussy

\paragraph{Results and discussions} 
\autoref{fig:ablation} shows that the IVMs without applying our technical protocols fail to produce consistent scores across dimensions and data cardinality, whereas they succeed when satisfying the axioms. The results demonstrate the superiority of our adjusted IVMs and the adjustment process.

All three protocols substantially contribute to achieving the axioms. 
The scatterplots (a-d, f-i) depict how SMAPE along the pairs of the controlled factor (data-cardinality, dimensionality) is changed between IVMs and their variants in average, while the bar plots (e, j) show the average error reduction made by the protocols and their combinations. 
The bar graphs indicate that the data-cardinality invariance protocol (T1; D) alone works as expected; it reduces the error by about 70\% in the data-cardinality test (bar D in e) and about 30\% in the dimensionality test (bar D in j).
The range invariance protocol (T4; R) also plays an important role in making measures less sensitive to varying factors, reducing the errors by about 40\% and 55\% alone for the data-cardinality (bar R in e) and dimensionality (bar R in j) tests, respectively.
Unexpectedly, the shift invariance protocol (T2; S) alone does not reduce error; in both tests, there exists a case in which the error increases after applying T2 (b, $\alpha$; g, $\beta\gamma$). However, the error generally reduces when we apply the shift invariance protocol together with data-cardinality or range invariance protocols (DS, RS, DRS bars in e, j). 
We interpret that the shift invariance protocol alone does not make apparent enhancement as the exponential protocol (T4-a,b) amplifies errors that are not yet reduced by either data-cardinality or range invariance protocols; applying other protocols reduces such errors and thus reveals the benefit of the shift invariance protocol. 

We find that interplays exist not only for the shift invariance protocol and generally provide positive effects.
In the dimensionality test, for example, both the shift invariance protocol and the data-cardinality invariance protocol perform better when they are combined (DS) than alone (D, S). The combination of all protocols (DRS) (red dots in e-d and f-i; DRS bar in e, j) also consistently shows good performance in making measures less sensitive to the change of dimensionality and data-cardinality. 
Meanwhile, applying all \textit{possible} protocols (\IVMa; denoted as ADJ) does not show better performance than most of the other combinations. This is possibly because some IVMs do not fully benefit from the interplay of all protocols (table on the bottom right). For example, for $SC$, only the shift invariance protocol, which shows a negative effect without other protocols, is applied.
$SC_A$ thus fails to reduce errors, compared to $SC$ (i, $\omega$). 
Still, applying all possible protocols generally makes IVM less sensitive to cardinality and dimension (d, i), which confirms the importance of our overall adjustment procedures (i.e., protocols) and their underlying axioms. 
Exploring and interpreting the interplays between the protocols and specific IVMs in more detail would be interesting for future work.

\subsection{Across-Dataset Rank Correlation Analysis}

\label{sec:bdsa}

\noindent 
\textbf{Objectives and design.}
Five \IVMa ($CH_A$, $DI_A$, $\{II, XB\}_A$, $DB_A$, and $SC_A$) are assessed against competitors (IVMs and classifiers)  for estimating the CLM ranking of labeled datasets. We approximate a ground truth (GT) CLM ranking of labeled datasets using multiple clustering techniques. We then compare the rankings made by all competitors to the GT using Spearman's rank correlation.

\paragraph{Datasets} We collect 96 publicly available benchmark labeled datasets from various sources (e.g., UCI ML repository \cite{asuncion07uci} and Kaggle \cite{kaggle2022}), with diverse numbers of data points, class labels, cluster patterns (presumably), and dimensionality (refer to Appendix G).

\paragraph{Approximating the GT CLM} For the lack of definite GT clusters in multidimensional real datasets, we use the maximum EVM score achieved by nine various clustering techniques (see below) on a labeled dataset as an approximation of the GT CLM score for that dataset. These GT scores are used to obtain the GT-ranking of all data sets. 
This scheme relies on the fact that high EVM implies good CLM (\autoref{sec:intro}; \autoref{fig:clm} A and D).
We use Bayesian optimization \cite{snoek12neurips} to find the best hyperparameter setting for each clustering technique. 
We obtain the GT-ranking based on the following four EVMs: adjusted rand index (\texttt{arand}) \cite{santos09icann}, adjusted mutual information (\texttt{ami}) \cite{vinh10jmlr}, v-measure (\texttt{vm}) \cite{rosenberg07emnlp}, and normalized mutual information (\texttt{nmi}) \cite{strehl02jmlr} with geometric mean. 
We select these measures because they are ``normalized'' or ``adjusted'' so that their scores can be compared across datasets \cite{wu09kdd}, and also widely used in literature \cite{liu18tpami, xiong17tpami, zhang22tpami, chakraborty22tpami}. 
For clustering techniques, we use HDBSCAN \cite{campello13akddm}, DBSCAN \cite{schubert17tds},  $K$-Means \cite{hartigan79jstor, likas03pr}, $K$-Medoids  \cite{park09esa}, $X$-Means \cite{pelleg00icml}, Birch \cite{zhang96sigmod}, and single, average, and complete variants of Agglomerative clustering \cite{mullner11arxiv} (Appendix F).  

\paragraph{Competitors} We compare \IVMa{}s, IVMs, and classifiers, which are natural competitors in measuring CLM (\autoref{sec:relwork}), to the GT-ranking. For classifiers, we use Support Vector Machine (SVM), $k$-Nearest Neighbors ($k$NN), Multilayer Perceptron (MLP), Naive Bayesian Networks (NB), Random Forest (RF), Logistic Regression (LR), Linear Discriminant Analysis (LDA), following  Rodr\'iguez et al. \cite{rodriques18kbs}. 
\revise{
We also use XGBoost (XGB), an advanced classifier based on tree boosting \cite{chen16kdd}. 
We use XGBoost as it adapts well regardless of the datasets' format \cite{chen16kdd, bohacek24trustnlp}, thus being suitable to all the 96 datasets composed of tabular, image, and text datasets.
XGBoost also outperforms recent deep-learning-based models in classifying tabular datasets \cite{grin22nips}, a preponderant type among our datasets.
Finally, we test the ensemble of classifiers.}
We measure the classification score of a given labeled dataset using five-fold cross-validation and Bayesian optimization \cite{snoek12neurips} to ensure the fairness of the evaluation. The accuracy in predicting class labels is averaged over the five validation sets to get a proxy of the CLM score for that dataset. For the ensemble, we get the proxy as the highest accuracy score among the \revise{eight} classifiers for each dataset independently\cite{rodriques18kbs}.

\definecolor{lightred}{RGB}{247, 163, 180}
\definecolor{lightlightred}{RGB}{245, 208, 216}

\newcommand{\lred}{\cellcolor{lightred}}
\newcommand{\llred}{\cellcolor{lightlightred}}

\begin{table}[t]
    \centering
    \caption{
    \label{tab:between_rank}
    The results of across-dataset rank correlation analysis (\autoref{sec:bdsa}). The numbers are the rank correlations between the approximated ground truth CLM ranking based on nine clustering techniques and the estimated CLM ranking obtained by \IVMa, IVMs, and classifiers. }  
    \scalebox{1}{
    \begin{tabular}{crrrrr}
    \toprule 
   %
    & &  \multicolumn{4}{c}{\makebox[0pt]{Ground truth ranking made by EVMs}} \\     
    & & \texttt{ami} & \texttt{arand} & \texttt{vm} & \texttt{nmi} \\ 
    \midrule
    \multirow{8}{*}{\rotatebox[origin=c]{90}{Classifiers}}                                  & NB & .4126 & .5276 & .3157  & .3130\\  
                                & MLP & .4405 & .5386 & .3600  & .3761\\ 
                                 & LR & .4456 & .5382 & .3666  & .3873\\ 
                                 & XGB & .4543 &  .5247  &  .3373     & .3377 \\
& $k$NN & .4876 & .5810 & .3974 & .4094\\ 
                                 & RF & .4893 & .5741 & .3991 & .3889\\ 
                                     & LDA & .4999 &  .5726 & .3945 & .3606\\ 
                                 & SVM & .5427 & \llred .6235 & .4625 & .4827\\  
                                 & Ensemble & .5536 &  .6162 \llred & .4486   & .4531 \\ 
    \midrule
         \multirow{6}{*}{\rotatebox[origin=c]{90}{IVM}}  
          & $CH$             & .5923& \llred .6222 & .4487 & .3810\\
          & $DI$            & .4026    & .3534 & .5366 & .5979\\
          & $II$          & .5668  & .5957 & \llred .6086 & \llred .6454\\
          & $XB$   & \llred .6201 & \llred .7019 & .4934 & .4446\\
          & $DB$   & \llred .7091 & \llred .7513 &  .5719 & .5015\\ 
          & $SC$ & .5648 & \llred .6800 & .4549 &   .4208\\
    \midrule
           \multirow{5}{*}{\rotatebox[origin=c]{90}{IVM$_{A}$}} 
           & $CH_{A}$   & \lred $^{**}$.8714& \lred $^{**}$.8472 & \lred $^{***}$.8300  & \llred $^{***}$.7836 \\
           & $DI_{A} $& \llred .7293 & \llred .7177& \llred .7504& \llred .7427 \\
           & $\{II, XB\}_{A}$& \lred $^{*}$.8463& \lred $^{*}$.8442& \lred $^{*}$.8060 & \llred $^{**}$.7818 \\
           & $DB_{A}$ & \lred .8315 & \lred .8111 & \llred .7856 & \llred .7436 \\ 
           & $SC_{A}$ & \lred$^{***}$.8955 & \lred $^{***}$.8769 & \lred $^{**}$.8217& \llred $^{*}$.7733 \\
 
    \bottomrule
    \addlinespace[0.115cm]
    \multicolumn{6}{l}{
        \footnotesize 
        \makecell[l]{ 
            (1) Every result was validated to be statistically significant  \\
            \hspace{3.3mm} ($p < .001)$ by Spearman's rank correlation test.
        }
    } \\
    \multicolumn{6}{l}{
        \footnotesize
        (2) $^{***}$ / $^{**}$ / $^{*}$: 1st- / 2nd- / 3rd-highest scores for each EVM
    } \\ 
    \multicolumn{6}{l}{\footnotesize
        (3) \textcolor{lightred}{$\blacksquare$} / \textcolor{lightlightred}{$\blacksquare$}: very strong ($>0.8)$ / strong ($>0.6)$ correlation \cite{prion14csn}
    } 
    \end{tabular}
    }
    
\end{table}

\paragraph{Results and discussions} 
\autoref{tab:between_rank} shows that for every EVM, \IVMa{}s outperform the competitors; first (***), second (**), and third (*) places are all part of the \IVMa category. \IVMa{}s
achieve about 17\% ($DB$) to 81\% ($DI$) of performance improvement, compared with their corresponding IVMs (average: 48\%), and have \textit{strong} (light-red cells) or \textit{very strong} (red cells) correlation with GT-ranking according to Prion et al.'s criteria \cite{prion14csn}. These results show that the adjustment procedure (T1-T4) relying on the new axioms (A1-A4), is beneficial to all IVMs, systematically increasing their correlation with the GT-ranking. Hence, \IVMa{}s are the most suitable measure to compare and rank datasets based on their CLM. Within the \IVMa{}s, $CH_A$ and $SC_A$ show the best performances, with a slight advantage for $CH_A$ (first place for both \texttt{vm} and \texttt{nmi}, and runner-up for both \texttt{ami} and \texttt{arand}). 

\begin{figure}
    \centering
    \includegraphics[width=\linewidth]{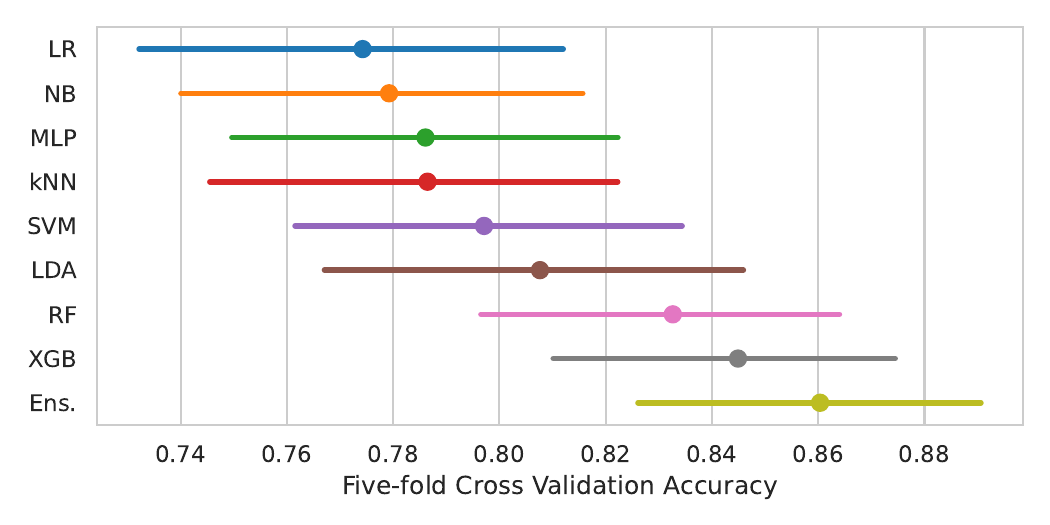}
    \caption{\revise{The cross-validation accuracy of classifiers in classifying classes in 96 labeled datasets. The classifiers are ordered based on the mean accuracy. Error bars indicate 95\% confidence interval. }}
    \label{fig:classifiers}
\end{figure}

In contrast, as expected, supervised classifiers \revise{fall behind the IVMs and \IVMa{}s, indicating that they} should not be relied upon for \revise{predicting CLMs. 
A notable finding is the most advanced model, XGB, shows relatively poor performance in estimating CLM compared to classical models such as SVM, $k$NN, and LDA; even an ensemble of classifiers falls behind SVM in terms of \texttt{arand}, \texttt{vm}, and \texttt{nmi} (\autoref{tab:between_rank}). 
This is because XGB and ensemble classifiers effectively discriminate classes regardless of whether they are well-separated by a large margin or not in the data space (\autoref{fig:classifiers} gray and yellow points), leading them to classify most datasets as having similarly good CLM. 
This finding indicates that improving classification accuracy does not necessarily help achieve better CLM measurement, further emphasizing the significance of our contribution.
}

\begin{table}
    \centering
        \caption{The results of within-dataset rank correlation analysis (\autoref{sec:wdca}).
    We compared the pairs of rankings obtained by IVM and \IVMa, with the ground truth noisy labels ranking (NR) on the 96 collected datasets. 
    }
    \begin{tabular}{r|c|cc}
    \toprule
    IVM    & \textit{IVM} vs. \textit{\IVMa}  &  NR vs. \textit{IVM}  & NR vs. \textit{\IVMa} \\
    \midrule
    $CH$    & \lred $.848\pm.263$ & \lred $.876\pm.296$ & \lred $.879\pm.277$ \\
    $DI$ &        $.253\pm.429$ &        $.451\pm.335$ &        $.381\pm.727$ \\
    $II$   & \lred $.825\pm.296$ & \lred $.857\pm.284$ &  \lred $.881\pm.307$ \\
    $XB$   &\lred $.820\pm.308$  & \lred $.832\pm.341$ & \lred $.881\pm.307$ \\
    $DB$    & \lred $.855\pm.257$ & \lred $.884\pm.268$ &  \lred $.876\pm.294$ \\
    $SC$   &        $.515\pm.574$ &        $.530\pm.601$ &\lred $.878\pm.322$ \\
    \bottomrule
    \addlinespace[0.115cm]
    \multicolumn{4}{l}{\footnotesize
    \makecell[l]{ 
            \textcolor{lightred}{$\blacksquare$} / \textcolor{lightlightred}{$\blacksquare$} : very strong ($>0.8)$ / strong ($>0.6)$ \\ 
            \hspace{9mm} correlation \cite{prion14csn}
        }
        
    } 
    \end{tabular}

    \label{tab:within_rank}
\end{table}

\begin{figure*}[]
    \centering
    \includegraphics[width=\textwidth]{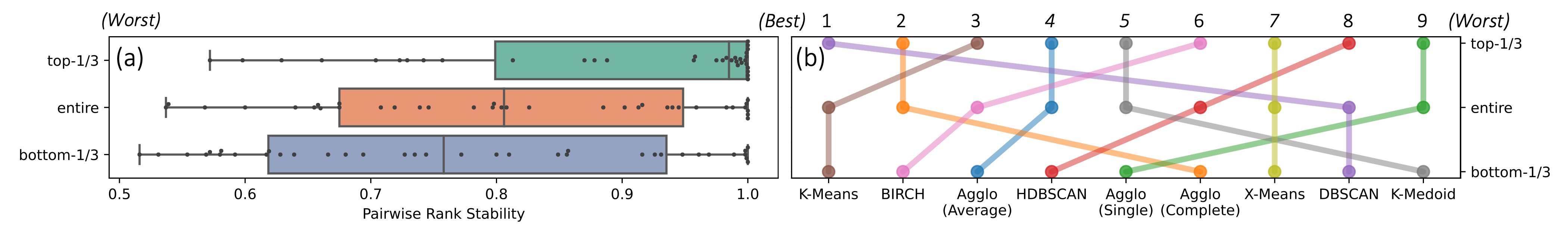}
    
     \vspace{-2.5mm}
    \caption{(a) 
     Distribution of pairwise rank stability for \texttt{bottom-1/3} (blue; $\mathcal{P}^-$), \texttt{entire} (orange; $\mathcal{P}^*$), and \texttt{top-1/3} (green; $\mathcal{P}^+$) subsets of 96 labeled datasets based on the CLM score computed by $CH_A$. (b) 
     Rankings of clustering techniques for each set; rankings are not stable and can change dramatically if we use low-quality datasets ((a) blue and orange bars). All rankings are based on \texttt{ami} averaged over each subset. Using the top-ranked datasets leads to more stable and reliable rankings ((a) green bar). } 
    \label{fig:rank_analysis}
\end{figure*}

\begin{figure}[]
    \centering
    \includegraphics[width=\linewidth]{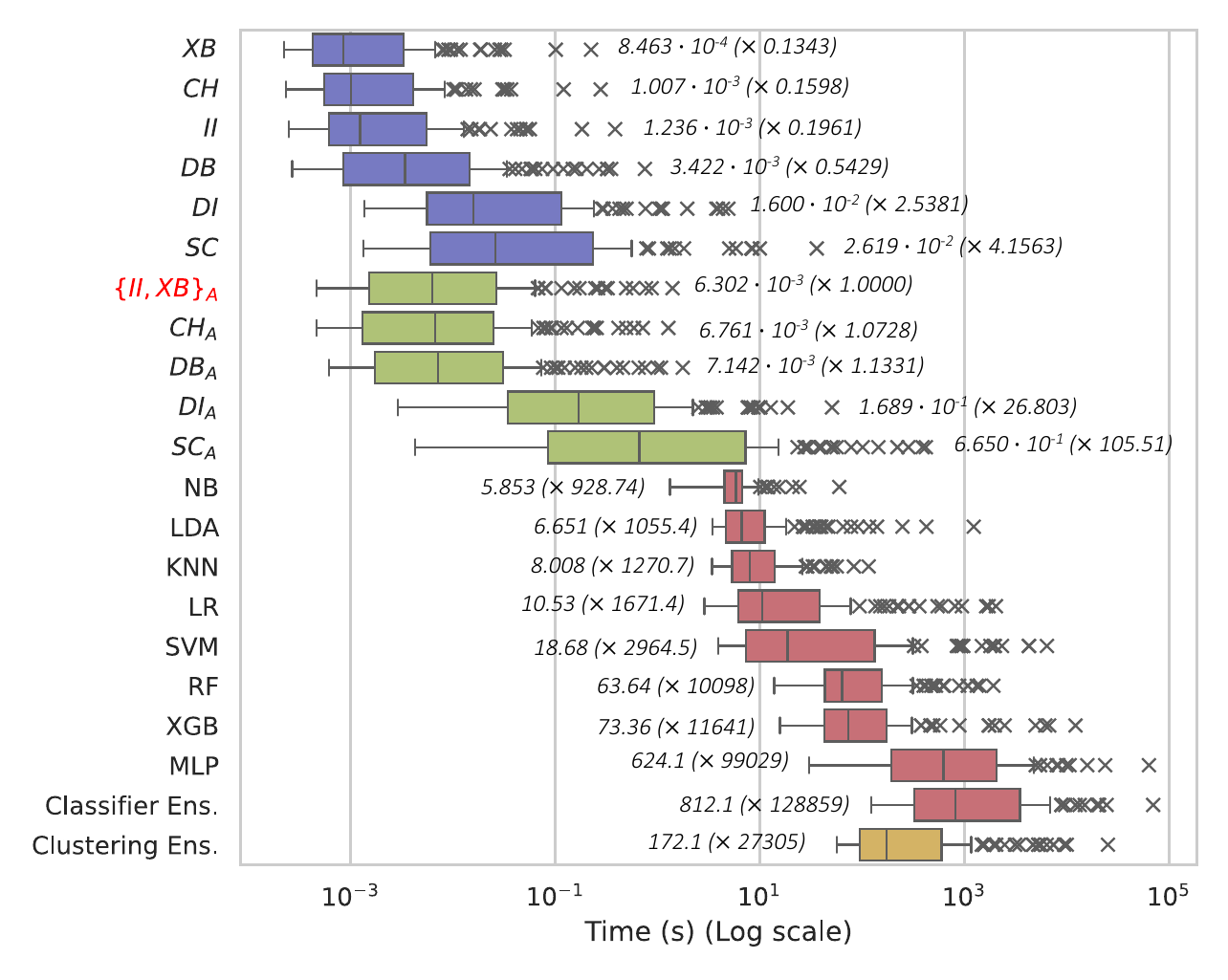}
    \vspace{-5mm}
    \caption{The runtime of the IVM (blue), \IVMa (green), classifiers (red), and the clustering ensemble (Ens.; yellow) in computing the CLM of 96 datasets. The numbers next to each box depict the median runtime of the corresponding measure (left) and the relative time compared to $\{II, XB\}_{A}$, the fastest \IVMa  (e.g., $CH_A$ is  1.07 times slower than $\{II, XB\}_{A}$).
    }
    \label{fig:runtime}
\end{figure}

\subsection{Within-Dataset Rank Correlation Analysis}

\label{sec:wdca}

\noindent

\noindent
\textbf{Objectives and design.}
We want to evaluate the \IVMa's ability to evaluate and compare CLM within a dataset, which is the original purpose of an IVM.
For this purpose, we generate several noisy label variants of each dataset and compare how the scores achieved by IVMs and their adjusted counterparts are correlated with the ground truth noisy label ranking (NR).
Assume a set of datasets $\mathcal{X} = \{X_i | i = 1, \cdots, n\}$ and their corresponding labels $\{C_j | j = 1, \cdots, n\}$. 
For each dataset $(X_k, C_k)$, we run the following process.
First, we generate 11 noisy label variants of each dataset $\{C_{k,l}| l = 0, \cdots, 10\}$ by randomly shuffling $l \cdot 10$\% of their labels.
The $l$-th noisy label dataset is authoritatively ranked at the $(11-l)$-th place of the NR (i.e., the larger the proportion $l$ of shuffled labels is, the lower the expected CLM). 
Then, for each IVM $Z \in \{CH, DI, II, XB, DB, SC\}$ and its corresponding \IVMa (i.e., $Z_A$), we compute the CLM ranking of these noisy label datasets based on 
$Z(C_{k, l}, X_k, \delta)$ and $Z_A(C_{k, l}, X_k, \delta)$, respectively.
We examine how 
\revise{the ranking generated by IVMs and \IVMa{}s}
are similar to NR using Spearman's rank correlation. 
We also check the rank correlation between \revise{the rankings from IVMs and \IVMa{}s}.

\paragraph{Datasets}
For $\mathcal{X}$, we use the 96 labeled datasets from \autoref{sec:bdsa}.

\paragraph{Results and discussions}
As shown in \autoref{tab:within_rank},
every \IVMa has a very strong rank correlation ($> 0.8$) with both NR and IVM for every case except for $DI_A$. 
The \IVMa{}s showed equal ($CH$, $DI$, $II$, $XB$, $DB$) or better ($SC$) performance in estimating the CLM within a dataset.
We also see that the discrepancy between IVM and \IVMa rankings follows the one between IVM and GT noisy labels ranking.
Such results verify the effectiveness of our protocols and \IVMa{}s in precisely measuring CLM within a dataset.

\subsection{Runtime Analysis}

\label{sec:runtime}

\noindent
\textbf{\revise{Objectives and design.}}
We compare the runtime of the approaches explored in \autoref{sec:bdsa}  to estimate the CLM of 96 labeled datasets.
For classifiers and the clustering ensemble, we measure the time of the entire optimization (\autoref{sec:bdsa}). 
See Appendix F for experimental settings, including the apparatus we use.

\paragraph{\revise{Results and discussion}}
As a result (\autoref{fig:runtime}), \IVMa{}s are up to one order of magnitude slower than $XB$, the fastest IVM. 
However, they are up to \textit{four orders of magnitude} ($\times 10,000$) faster than the competitors like clustering ensemble used to estimate ground truth CLM in \autoref{sec:bdsa}. This verifies that most \IVMa{}s, among which is $CH_A$, show an excellent tradeoff between accuracy and speed. Despite $SC_A$ being as accurate as $CH_A$ (\autoref{sec:bdsa}), it is \textit{two orders of magnitude} slower ($\times 0.01$), making $CH_A$ the best \IVMa to use in practice.

\begin{figure}[t]
    \centering
    \includegraphics[width=\linewidth]{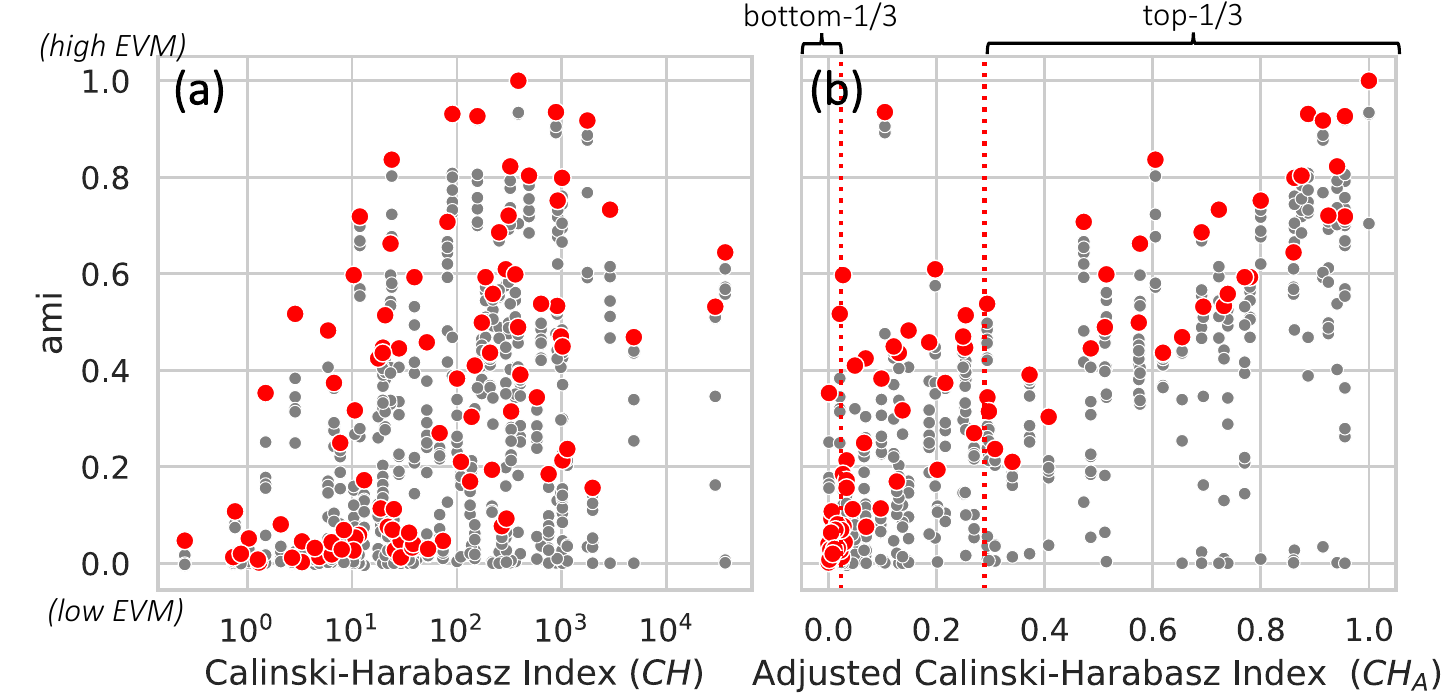}
    \caption{
    All (gray points) and best (red points) \texttt{ami} scores of GT clustering techniques for the $96$ benchmark datasets ranked by $CH$ (left) and $CH_A$ (right). 
    The correlation with ground truth is clearly better for the adjusted index (\autoref{tab:between_rank}).
    The \texttt{top-1/3} datasets (\pplus) in terms of $CH_A$ (right) are the most reliable datasets to use to evaluate and compare clustering techniques using EVMs (\autoref{sec:selection}). 
    }
    \label{fig:btw_result}
\end{figure}

\section{Applications}

\label{sec:applications}

\revise{
We present two applications of \IVMa{}s. First, we show that evaluating the CLM of benchmark datasets beforehand and using only those with the highest CLM scores enhances the stability and robustness of external validation and ranking of clustering techniques (\autoref{sec:selection}). We also show that \IVMa{}s can be leveraged to improve the CLM of benchmark datasets (\autoref{sec:improvclm}). 
}

\subsection{Ranking Benchmark Datasets for Reliable EVM}

\label{sec:selection}

\noindent
\textbf{Objectives and design.} We want to demonstrate the importance of evaluating the CLM of benchmark datasets prior to conducting the external validation. 
Here, we consider the \texttt{entire} set of $96$ labeled 
datasets ($\mathcal{P}^*$), and also the \texttt{top-1/3}  ($\mathcal{P}^+$) and  \texttt{bottom-1/3}  ($\mathcal{P}^-$) datasets (\autoref{fig:btw_result}) based on their $CH_A$ ranking.
We consider simulating the situation whereby a data scientist would arbitrarily choose $10$ benchmark datasets ($\mathcal{B}$) among the datasets at hand for the task $T$ of\textit{ ranking clustering techniques according to $EVM_{\mathcal{B}}$}, the average EVM over $\mathcal{B}$.
For each $\mathcal{P}\in\{\mathcal{P}^+,\mathcal{P}^*,\mathcal{P}^-\}$, we simulate $100$ times picking $\mathcal{B}$ at random among $\mathcal{P}$. For each $\mathcal{P}$, we measure the pairwise rank stability $P_\mathcal{B}(A,B)=\max(1-p,p)$ of clustering techniques A and B over $\mathcal{B}$ by counting the proportion $p$ of cases $\texttt{ami}_{\mathcal{B}}(A)>\texttt{ami}_\mathcal{B}(B)$.

\sloppy
\paragraph{Hypothesis} We expect that conducting $T$ on any subset of good-CLM datasets will provide similar rankings (\autoref{fig:clm}A), where pairwise ranks remain stable ($\forall (A,B), P_\mathcal{B}(A,B)\approx 1$), whereas conducting $T$ using bad-CLM datasets will lead to arbitrary and unstable rankings ($\forall (A,B), P_\mathcal{B}(A,B)$ spreads over $[0.5, 1]$) 
(\autoref{fig:clm}BEH). 

\fussy

\paragraph{Results and discussion} As depicted in \autoref{fig:rank_analysis}a, picking datasets from \pplus{}  provides more stable rankings of clustering techniques, compared with \pminus{} and \pentire, which validates our hypothesis. 
Moreover, we find that the rankings of techniques made by $\texttt{ami}_{\mathcal{P^+}}$, $\texttt{ami}_{\mathcal{P^*}}$, and $\texttt{ami}_{\mathcal{P^-}}$ are drastically different (\autoref{fig:rank_analysis}b; e.g., DBSCAN is in first place with $\mathcal{P}^+$, but in eighth place with $\mathcal{P}^*$ or $\mathcal{P}^-$). 
Still, some datasets within $\mathcal{P}^-$ (e.g., Spambase and Hepatitis \cite{asuncion07uci}) 
were used for external validation in previous studies \cite{khan21icecit, monath19kdd}
without CLM evaluation, casting doubt on their conclusions and showing that this issue shall gain more attention in the clustering community.

\subsection{Improving the CLM of Benchmark Datasets}

\label{sec:improvclm}

\noindent
\textbf{Objectives and design.}
While datasets with good CLM lead to more reliable EVM (\autoref{sec:selection}), there is a limited number of such datasets. We thus propose to improve the CLM of existing datasets to enhance diversity and robustness in executing EVM. As a proof-of-concept, we show that \IVMa{}s can be used to improve the CLM of labeled datasets by implementing a feature selection algorithm that finds a subspace of a given dataset that maximizes \IVMa{} scores.
Formally, given $X$, $C$, $\delta$, and \IVMa $f$, the algorithm finds the binary weight vector $w^*\in \{0,1\}^{|X|}$ that satisfies:
\begin{equation}
    w^* = \argmax_{w\in\{0,1\}^{|X|}} f(C, w \cdot X,\delta),
\end{equation}
where $\cdot$ denotes the column-wise weight product: 
\revise{
\begin{equation}
    w \cdot X = (w^T * \mathbf{1}) \odot X,
\end{equation}
}
where $\odot$ denotes the Hadamard product and $*$ represents a standard matrix product. 
The algorithm returns the column-filtered data $X^* = w^* \cdot X$ as output. 
We generate 1,000 random weight vectors $w$ and pick the one that maximizes $f$, while using $CH_A$ as $f$.
We run the algorithm for each of the 96 labeled datasets  ($\mathcal{P}^*$) and compare the CLM of the original datasets to their optimally-filtered counterparts. 
We also repeat the rank stability experiment (\autoref{sec:selection}) using the improved datasets. We evaluate how the improved set differs from the original set in terms of rank stability. Finally, we record the runtime of the algorithm to check whether the improvement is achieved in a reasonable time.

\begin{figure}
    \centering
    \includegraphics[width=\linewidth]{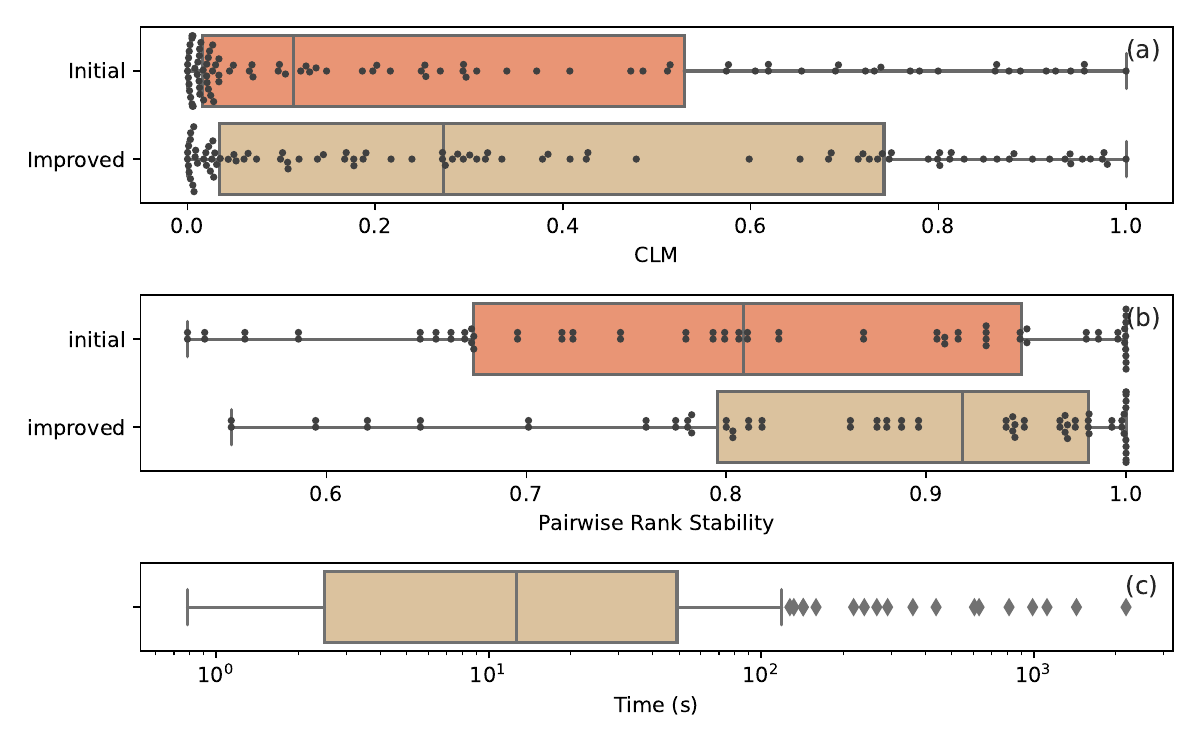}
    \vspace{-8mm}
    \caption{(a) A comparison of the CLM of 96 labeled datasets before (initial) and after (improved) applying our algorithm that uses $CH_A$ to enhance CLM (\autoref{sec:improvclm}). 
    (b) A comparison of rank stability of EVM using the initial and improved dataset.
    (c) The runtime of the algorithm. 
    The results confirm that \IVMa{}s can enhance the reliability of EVM benchmark datasets within a reasonable time, lower than a minute for most datasets.
    }
    \label{fig:improvement}
\end{figure}

\paragraph{Results and discussions}
As illustrated in  \autoref{fig:improvement}a, there is a substantial overall increase in the CLM of datasets, which confirms the effectiveness of \IVMa{}s in improving CLM. 
The improved dataset also outperformed the original datasets in terms of rank stability (\autoref{fig:improvement}b), which shows that the improved dataset substantially enhances the reliability of EVM.
We also find that the algorithm completes in less than 20 seconds on average (\autoref{fig:improvement}c), and takes less than an hour even for the largest dataset. 
In summary, utilizing \IVMa allows for readily generating reliable benchmark datasets for EVM within a practical time frame.

\section{Discussions}

\subsection{Benefits of Adjusted IVMs}

Utilizing an \IVMa{} to estimate the CLM of a dataset drastically reduces the runtime by \textit{four orders of magnitude} compared to the clustering ensemble (\autoref{sec:runtime}); this translates to a one-year CLM computation being shortened to just 53 minutes.
Given this computational efficiency, \IVMa{}s can be used to improve the CLM of labeled datasets by selecting their dimensions, even with a naive and costly random search (\autoref{sec:improvclm}).
Such improvement enables applications that require real-time measurement of CLM. For example, it is now possible to estimate the CLM quality on the fly for streaming data 
without resorting to distributed computing or approximate solutions. We can also compute the CLM of very large datasets, e.g., image or text data in their transformer \cite{vaswani17neurips} space during the training phase, to evaluate the quality of this representation at each iteration or in each layer\cite{dalvi2022discovering}, yet a task impracticable with clustering ensembles.

\subsection{Consistency, Soundness, and Completeness of the Axioms}    

\label{app:axioms_discussion}

We discuss the consistency, soundness, and completeness of the extended axioms (within- and across-dataset axioms), following \AB \cite{bendavid08nips}.
We remind the definition of these three characteristics:
\textbf{(1) Consistency:} the set of axioms is consistent if at least one ideal object (i.e., \IVMa) satisfies all of them, which means that there is no contradiction between the axioms.
We use the same terminology with W2 (consistency of IVM), but with a different meaning (consistency of axioms). 
\textbf{(2) Soundness:} the set of axioms is sound if every object existing in the target group (i.e., every \IVMa) satisfies all the axioms. 
\textbf{(3) Completeness:} the set of axioms is complete if any object not included in the target group (i.e., non-\IVMa) fails at least one axiom. 
We discuss them for \IVMa{s}:

\paragraph{Consistency} The existence of five \IVMa{s} that satisfy all within- and across-dataset axioms (Appendix D) validates their consistency.

\paragraph{Soundness}
Soundness cannot be proven as we lack a clear definition that differentiates \IVMa{}s and non-\IVMa{}s.
However, if we define \IVMa as the set of measures that can be adjusted from any IVM, the axioms are sound as our technical protocols guarantee that all adjusted measures satisfy or approach the axioms by design. 
We suggest both defining a valid \IVMa as a ``function that satisfies all within-dataset and across-dataset axioms'' and using the axioms as guidelines to adjust other IVMs or to design new \IVMa, thereby preserving the soundness of all the axioms.

\paragraph{Completeness}
Our extended axioms are more complete than the within-dataset axioms in terms of defining \IVMa, given that standard IVMs do not meet the new across-dataset axioms. 
The across-dataset rank correlation analysis (\autoref{sec:bdsa}) empirically quantifies and validates this increased completeness.  
However, it remains uncertain if our axioms cover every aspect that can vary across datasets.
There might be a function that fulfills all axioms but is not a valid \IVMa. Searching for such a function and making the axioms more complete will be an interesting avenue for future work. 
\section{Conclusion}

In this research, we provide a grounded way to evaluate the validity of labeled datasets used as benchmarks for external clustering validation. We propose doing so by measuring their level of cluster-label matching (CLM). 
We propose new across-dataset axioms and technical protocols to generate measures that satisfy the axioms. We use these protocols to design five adjusted internal validation measures (\IVMa{}s), generalizing standard IVMs, to estimate the CLM of benchmark labeled datasets and rank them accordingly. 
\revise{A series of experiments confirm \IVMa{}s' accuracy, scalability, and practical effectiveness in supporting reliable external clustering validation.} 
As the primary practical outcome of this work, the $96$ datasets ranked by CLM estimated by the proposed \IVMa{}s measures are available at \href{https://github.com/hj-n/labeled-datasets}{github.com/hj-n/labeled-datasets} for use by practitioners to generate more trustworthy comparative evaluations of clustering techniques.

As future work, we would like to explore other uses of \IVMa and design new axioms to build better clustering benchmarks. 
For example, designing \IVMa{}s that consider non-globular and hierarchical clusters would be an interesting path to explore.
Developing better optimization techniques for maximizing the CLM of benchmark datasets is another path worthy of further study. Finally, given the high scalability of \IVMa{}s, we envision using these CLM measures to compare cluster structures of labeled data across different latent spaces, like in the layers of foundational models for exploring concept representations during pre-training or fine-tuning.

The datasets and code are available at \href{https://github.com/hj-n/labeled-datasets}{github.com/hj-n/labeled-datasets} and \href{https://github.com/hj-n/clm}{github.com/hj-n /clm}, respectively.

\section*{Acknowledgments}

This work was supported by the National Research Foundation of Korea (NRF) grant funded by the Korean government (MSIT)  (No. 2023R1A2C200520911) and the Institute of Information \& communications Technology Planning \& Evaluation (IITP) grant funded by the Korean government (MSIT) [NO.RS-2021-II211343, Artificial Intelligence Graduate School Program (Seoul National University)]. 
The ICT at Seoul National University provided research facilities for
this study.
Hyeon Jeon is in part supported by Google Ph.D. Fellowship.

\bibliographystyle{IEEEtranDOI}
\bibliography{ref}

\vfill

\end{document}